\pdfoutput=1

\documentclass[11pt]{article}

\usepackage[]{acl}

\usepackage{times}
\usepackage{latexsym}
\usepackage{float}
\usepackage[T1]{fontenc}

\usepackage[utf8]{inputenc}

\usepackage{graphicx}
\usepackage{microtype}
\usepackage{amsfonts}
\usepackage{mathrsfs}
\usepackage{multirow}
\usepackage{amssymb}
\usepackage{subfigure} 
\usepackage{CJKutf8}
\usepackage{colortbl}
\usepackage{color}
\usepackage[table]{}
\usepackage{booktabs}
\usepackage{makecell}

\title{That Slepen Al the Nyght with Open Ye! Cross-era Sequence Segmentation with Switch-memory}

\author{Xuemei Tang$^{1,2}$  Qi Su$^{2,3,4}$\thanks{\quad Corresponding author}  Jun Wang$^{1,2}$
\\$^1$Department of Information Management, Peking University, Bejing, China
\\$^2$Digital Humanities Center of Peking University, Bejing, China
\\ $^3$School of Foreign Languages, Peking University, Bejing, China
\\ $^4$MOE Key Lab of Computational Linguistics, School of EECS, Peking University, Bejing, China
\\
\texttt{tangxuemei@stu.pku.edu.cn}  \\\texttt{\{sukia,junwang\}@pku.edu.cn}}

\begin{document}
\begin{CJK*}{UTF8}{gbsn}
\maketitle
\begin{abstract}
The evolution of language follows the rule of gradual change. Grammar, vocabulary, and lexical semantic shifts take place over time, resulting in a diachronic linguistic gap. As such, a considerable amount of texts are written in languages of different eras, which creates obstacles for natural language processing tasks, such as word segmentation and machine translation. Although the Chinese language has a long history, previous Chinese natural language processing research has primarily focused on tasks within a specific era. Therefore, we propose a cross-era learning framework for Chinese word segmentation (CWS), CROSSWISE, which uses the Switch-memory (SM) module to incorporate era-specific linguistic knowledge. Experiments on four corpora from different eras show that the performance of each corpus significantly improves. Further analyses also demonstrate that the SM can effectively integrate the knowledge of the eras into the neural network.
\end{abstract}

\section{Introduction}

As a human-learnable communication system, language does not remain static but instead evolves over time. The rate of change between different aspects of language, such as grammar, vocabulary, and word meaning, vary due to language contact and many other factors, which has led to the diachronic linguistic gap. An example of this can be seen in, ``That slepen al the nyght with open ye (That sleep all the night with open eye),'' which is a sentence from The Canterbury Tales, written in Middle English by Geoffrey Chaucer at the end of the 14th century. It is difficult for people without an understanding of Middle English to make sense of this sentence. Furthermore, some discourses contain both modern English and Old English due to citation or rhetorical need. For example, Shakespeare's fourteen lines of poetry are often quoted in contemporary novels. This kind of era-hybrid text creates barriers to natural language processing tasks, such as word segmentation and machine translation.

\begin{table}[t]
\centering
\setlength{\tabcolsep}{0.00001mm}{
\begin{tabular}{|l|c|c|c|c|c|c|c|c|} 
\hline
\rowcolor{gray!25}
\multicolumn{9}{|l|}{Sample from MSR}                                                                                         \\ 
\hline
\multirow{2}{*}{Golds} & \multicolumn{2}{c|}{(wait)} & (who)   & (come)        & \multicolumn{2}{c|}{(slove)}   & (ne)  & ？  \\ 
\cline{2-9}
& \multicolumn{2}{c|}{等待}     & 谁       & 来             & \multicolumn{2}{c|}{解决} & 呢 & ？  \\ 
                       
\hline
PKUSeg                 & \multicolumn{2}{c|}{等待}     & 谁       & 来             &  \multicolumn{2}{c|}{解决}       & 呢     & ？  \\ 
\hline
JiaYan                 & 等    & 待                    & 谁       & 来             & 解                 & 决            & 呢     & ？  \\ 
\hline
\rowcolor{gray!25}
\multicolumn{9}{|l|}{Sample~ from AWIKI}                                                                                       \\ 
\hline
\multirow{2}{*}{Golds} & (Qi) & \multicolumn{2}{c|}{(Cui Shu)} & \multicolumn{2}{c|}{(lead army)} & (attack)     & (Lv)  & 。  \\  
\cline{2-9}
                       
                       & 齐    & \multicolumn{2}{c|}{崔杼}        & \multicolumn{2}{c|}{帅师}           & 伐            & 莒     & 。  \\ 
\hline
PKUSeg                 & \multicolumn{2}{c|}{齐崔}     & \multicolumn{2}{c|}{杼帅} & \multicolumn{3}{c|}{师伐莒}                 & 。  \\ 
\hline
JiaYan                 & 齐    &  \multicolumn{2}{c|}{崔杼}                    & \multicolumn{2}{c|}{帅师 }                     & 伐         & 莒     & 。  \\
\hline
\end{tabular}}
\caption{Illustration of the different segmentation results for a modern Chinese sentence and an ancient Chinese sentence with different segmentation toolkits.}
\label{t1}
\end{table}

The Chinese language has the honor of being listed as one of the world’s oldest languages and, as such, has seen several changes over its long history. It has undergone various incarnations, which are recognized as Archaic (Ancient) Chinese, Middle Ancient Chinese, Near Ancient Chinese, and Modern Chinese. Notably, most Chinese NLP tasks skew towards Modern Chinese. Previous research has primarily focused on addressing the CWS problem in Modern Chinese and has achieved promising results, such as Chinese Word Segmentation (CWS) \citep{em:7,em:8,em:9,em:20,em:17,em:36, em:35,em:30,em:31}. Although CWS for ancient Chinese has been recognized in recent years, the processing of language-hybrid texts is still an open question. As shown in Table~\ref{t1}, PKUSeg \citep{em:38} is a Chinese segmenter that is trained with a modern Chinese corpus; while it can segment modern Chinese sentences correctly, its accuracy drops sharply when applied to ancient Chinese. Conversely, the ancient Chinese segmenter JiaYan\footnote{\url{http://github.com/jiayan/Jiayan/}} performs well on ancient Chinese text but fails to perform well on Modern Chinese texts. Therefore, it is necessary to develop appropriate models to undertake cross-era NLP tasks. 

To address this need, we propose CROSSWISE (CROsS-ear Segmentation WIth Switch-mEmory), which is a learning framework that deals with cross-era Chinese word segmentation (CECWS) tasks. The framework integrates era-specific knowledge with the Switch-memory mechanism to improve CWS for era-hybrid texts. More specifically, we utilized the abilities of both CWS and sentence classification tasks to predict segmentation results and era labels. We also incorporated the Switch-memory module to include knowledge of different eras, which consists of key-value memory networks \cite{em:19} and a switcher. In order to store era-specific knowledge by several memory cells, key-value memory networks are used. The sentence discriminator is considered to be a switcher that governs the quantity of information in each memory cell that is integrated into the model. For each memory cell, we map candidate words from the dictionary and word boundary information to keys and values..

The main contributions of this paper are summarized as follows:
\begin{itemize}
\item Cross-era learning is introduced for CWS, we share all the parameters with a multi-task architecture. The shared encoder is used to capture information that several datasets from different eras have in common. This single model can produce different words segmentation granularity according to different eras.
\item The Switch-memory mechanism is used to integrate era-specific knowledge into the neural network, which can help improve the performance of out of vocabulary (OOV) words. This study proposes two switcher modes (\emph{hard-switcher} and \emph{soft-switcher}) to control the quantity of information that each cell will feed into the model. 
\item Experimental results from four CWS datasets with different eras confirm that the performance of each corpus improves significantly. Further analyses also demonstrate that this model is flexible for cross-era Chinese word segmentation.
\end{itemize}

\begin{figure*}[t]
\centering
\includegraphics[height=4.8cm,width=16.5cm]{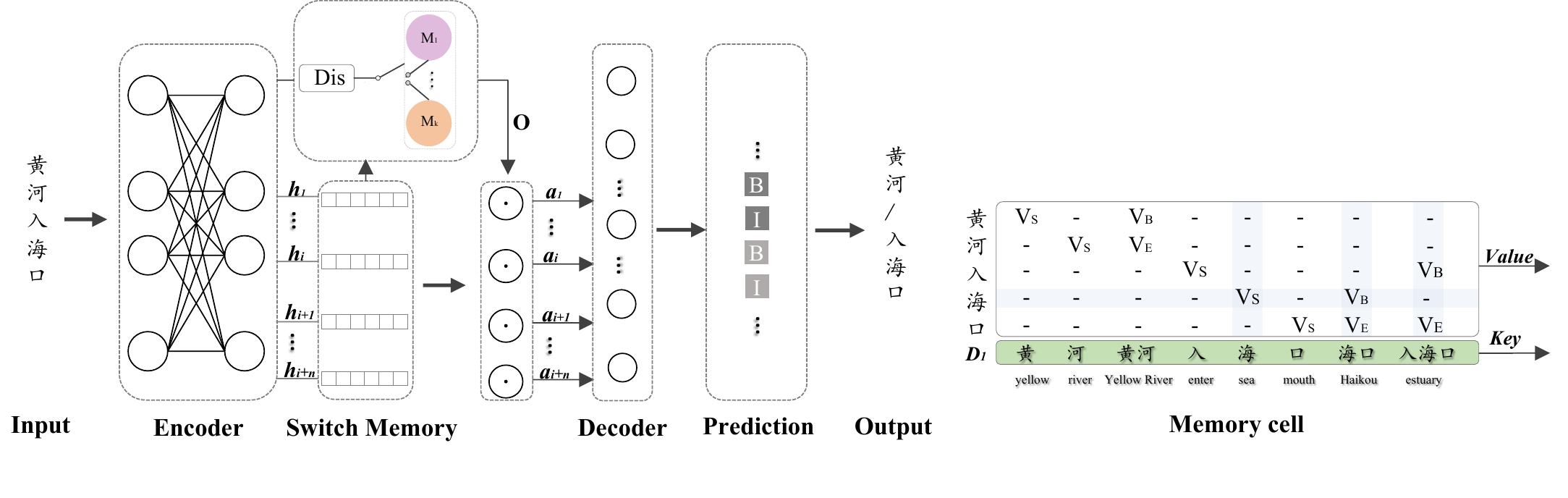}

\caption{CROSSWISE for cross-era Chinese word segmentation. ``Dis'' represents the discriminator, specially, the sentence classifier. ``$\rm M_1$'' is the first memory cell; its internal structure is shown on the right of the figure. For each character, the first memory cell extracts all candidate words from the input sentence and only retains ones that appeared in the first dictionary (candidate words as keys, words' boundary information as value). }
\label{p1}
\end{figure*}

\section{Related Work}

Chinese word segmentation is generally considered to be a sequence labeling task, namely, to assign a label to each character in a given sentence. In recent years, many deep learning methods have been successfully applied to CWS \citep{em:7,em:8,em:9,em:20,em:17,em:36,em:37,em:34,em:35,em:27,em:29,em:42,em:30,em:31,em:26,em:24}. Among these studies, some indicate that context features and external knowledge can improve CWS accuracy \citep{em:37,em:36,em:35,em:34,em:30,em:31,em:26}. Studies from \citet{em:34} and \citet{em:35} leveraged the dictionary to improve the task; n-gram is also an effective context feature for CWS \citep{em:37,em:30,em:17}.  The use of syntactic knowledge generated by existing NLP toolkits to improve CWS and part-of-speech (POS) has been established by \citet{em:30}. Furthermore, \citet{em:26} incorporated wordwood information for neural segmenters and achieved a state-of-the-art performance at that time.

It is common practice to jointly train CWS and other related tasks based on a multi-task framework. \citet{em:4} took each segmentation criterion as a single task and proposed an adversarial multi-task learning framework for multi-criteria CWS by extracting shared knowledge from multiple segmentation datasets. \citet{em:36} investigated the effectiveness of several external sources for CWS by a globally optimized beam-search model. They considered each type of external resource to be an auxiliary classification, and then leveraged multi-task learning to pre-train the shared parameters used for the context modeling of Chinese characters. \citet{em:34} jointly trained the CWS and word classification task by a unified framework model. Inspired by these successful studies, this study also incorporated ideas from the multi-task framework, and jointly trained the CWS task and the sentence classification task to enhance the performance of cross-era CWS.

Recently, some studies have noticed the linguistic gap due to the differences between eras. \citet{em:45} proposed a time-aware re-contextualization approach to bridge the temporal context gap. \citet{em:46} reframed the translation of ancient Chinese texts as a multi-label prediction task, then predicted both translation and its particular era by dividing ancient Chinese into three periods.

The use of key-value memory networks were introduced to the task of directly reading documents and answering questions by \citet{em:19}, which helped to bridge the gap between direct methods and the use of human-annotated or automatically constructed Knowledge Bases. \citet{em:26} applied this mechanism to incorporate n-grams into the neural model for CWS.

Encouraged by the above works, this study designed a multi-task model for cross-era CWS by jointly training the sentence classification task and CWS through the use of a unified framework model. Key-value memory networks are used to integrate era-specific knowledge into the neural network, as was done in research by \citet{em:26}.

\section{The Proposed Framework}

\subsection{BERT-CRF model for Chinese word Segmentation}
Chinese word segmentation is generally viewed as a character-based sequence labeling task. Specifically, given the sentence $X\ = \{ x_1,x_2, … , x_T \}$, each character in the sequence is labeled as one of $\mathcal{L}\ = \{B,M,E,S\}$, indicating the location of the character as at the beginning, middle, or end of a word, or that the character is a single-character word. CWS aims to determine the ground truth of labels  $ Y^* \ = \{ y_1^*, y_2^*, … , y_{T}^* \} $:

\begin{equation}\label{eq1}
Y^\ast={arg\ max\ P\left(Y\middle| X\right)}_{Y\in \mathcal{L}^T} 
\end{equation}

The universal end-to-end neural CWS architecture usually contains an encoder and a decoder. The framework used in this study is shown in Figure~\ref{p1}; the functions of each part are explained below.

\textbf{Encoding layer.} According to \citet{em:39}, although BERT-based \citep{em:47} models for CWS are imperfect, BERT is superior, in many aspects, to models that have not been pre-trained. For example, BERT is more suitable for dealing with long sentences; therefore, this study utilizes BERT released by Google \citet{em:47} as the shared encoder, which is pre-trained with a large amount of unlabeled Chinese data.

\begin{equation}\label{eq2}
\{\textbf{h}_1...\textbf{h}_i...{\textbf{h}}_T\} = Encoder(\{x_1...x_i...x_{T}\})
\end{equation}
where $\textbf{h}_i$ is the representation for $x_i$ from the encoder.

\textbf{Decoding layer.} This study is able to use a shared decoder for samples from different eras because era-aware representation have been combined for each character by the Switch-memory module. There are various algorithms that can be implemented as decoders, such as conditional random fields (CRF) \citep{em:10} and softmax. According to \cite{em:26}, CRF performs better in word segmentation tasks. Therefore, considering the framework of this study, CRF is used as the decoder.



In the CRF layer, $P\left(Y\middle| X\right)$  in Eq. \ref{eq1} can be represented as:

\begin{equation}\label{eq3}
P(Y|X)=\frac{\emptyset(Y|X)}{\sum_{Y^\prime\in \mathcal{L}^T}{\emptyset(Y^\prime|X)}}
\end{equation}
where, $\ \ \emptyset(Y|X)$ is the potential function, and only interactions between two successive labels are considered.

\begin{equation}\label{eq4}
\emptyset(Y|X)=\ \prod_{i=2}^{T}{\sigma(X,i,y_{i-1},y_i)}
\end{equation}

\begin{equation}\label{eq5}
\sigma(\mathbf{x},i,y^\prime,y)\ =\ exp(s{(X,i)}_y+b_{y^\prime y})
\end{equation}
where $b_{y^\prime y} \in\mathbb{\rm \textbf{R}}$ is trainable parameters respective to label pair $(y^\prime,\ y)$. The score function $s(X\ ,\ i)\ \in\mathbb{R}^{\left|\mathcal{L}\right|}$ calculate the score of each lable for $i_{th}$ character:

\begin{equation}\label{eq6}
s(X\ , \ i)\ ={\ \textbf{W}}_s^\top \textbf{a}_i+b_s
\end{equation}
where $\textbf{a}_i$ is the final representation for $i_{th}$ character. $\textbf{W}_s\in\mathbb{R}^{d_a\times L}$ and $b_s\in\mathbb{R}^{\left|\mathcal{L}\right|}$ are trainable parameters.

\subsection{Switch-memory mechanism}
The Switch-memory consists of \emph{d} memory cells and a switcher. For an input sentence, there are \emph{d} memory cells for each character. The switcher governs how much information in each cell will be integrated into the network. And the state of the switcher depends on the sentence classifier.

\subsubsection{Memory cells} The dictionary has been a useful external source to improve the performance of CWS in many studies.\citep{em:36,em:34,em:35}. However, the ability to incorporate the dictionary into previous research has been limited by either concatenating candidate words and character embeddings or the requirement of handcrafted templates. In this study, key-value memory networks are utilized to incorporate dictionary information, which is initially applied to the Question Answering (QA) task for improved storage of prior knowledge required by QA. Furthermore, this network structure can also be used to store the existing knowledge that is required by cross-era CWS.

Ancient Chinese is not a static language but is instead a diachronic language. Ancient Chinese has three development stages: Ancient, Middle Ancient, and Near Ancient. Each stage has a specific lexicon and word segmentation granularity. Therefore, this research has constructed four dictionaries $\mathcal{D} = \{D_0, D_1, D_2, D_3\}$, associating with the four development stages of Chinese, respectively, and each dictionary is era-related. When input a sentence, four memory cells are generated for each character in the sentence according to the four dictionaries, and each memory cell maps candidate words and word boundary information to keys and values.

\begin{table}
\centering

\begin{tabular}{l|ll|c||} 
\hline
Rule                                        & $V_{i,j}$  \\ 
\hline
$x_i$ is the beginning character of $w_{i,j}$. & $V_B$    \\ 
\hline
$x_i$ is the ending character of  $w_{i,j}$.    & $V_E$    \\ 
\hline
$x_i$ is a single word,  $w_{i,j}$.             & $V_S$    \\
\hline
\end{tabular}
\caption{the rules for assigning different values to $x_i$ according to its position in word $w_{i,j}$.}
\label{t2}
\end{table}

\textbf{Candidate words as keys.} Following \citeauthor{em:26}, for each $x_i$ in the input sentence, each dictionary has many words containing $x_i$, we only keep the n-grams from  the input sentence and appear in each dictionary, resulting $w_i^d\ = \{w_{i,1}^d,\ w_{i,2}^d,...w_{i,j}^d...w_{i,m_i}^d\}$ , $x_i$ is a part of word $w_{i,j}^d\in D_d$, $d\in[0,3]$. We use an example to illustrate our idea. For the input sentence show in Figure \ref{p1}, there are many n-grams containing $x_3=\mbox{``海(sea)''}$, we only retain ones that appear in $D_0$ for the first memory cell, thus, $w_3^0=\{``\mbox{海口(HaiKou)''}, ``\mbox{入海口(estuary)''}, ``\mbox{海(sea)''}\}$. Similarly, we can generate $w_3^1$, $w_3^2$, $w_3^3$ for the second, third and fourth memory cell according to $D_1, D_2, D_3$. Then, the memory cell compute the probability for each key (which are denoted as ${\ e}_{i,j}^w$ for each ${\ w}_{i,j}^d$), here $\textbf{h}_i$ is the embedding for $x_i$, which is encoded by the encoder.

\begin{equation}\label{eq7}
p_{i,j}^d=\ \frac{exp(\textbf{h}_i\cdot{\ e}_{i,j}^w\ )}{\sum_{j=1}^{m_i}{exp(\textbf{h}_i\cdot{\ e}_{i,j}^w\ )}}\ 
\end{equation}

\textbf{Word boundary information as values.} As we know, CWS aims to find the best segment position. However, each character $x_i$ may have different positions in each ${\ w}_{i,j}^d$. For example, $x_i$ may be at the beginning, middle, ending of ${\ w}_{i,j}^d$, or $x_i$ maybe a single word. Different positions convey different information. Therefore, we use the boundary information of candidate words as values for key-value networks. As shown in Table \ref{t2}, a set of word boundary values $\{V_B, V_E, V_S\}$ with embeddings $\{e_{V_B}, e_{V_E}, e_{V_S}\}$ represent the $x_i$’s different positions in ${\ w}_{i,j}^d$, and we map $x_i$ to different value vectors according to its positions. As a result, each $w_i^d$ for $x_i$ has a values list $\mathcal{V}_i^d\ = \ [v_{i,1}^d,v_{i,2}^d,{...v}_{i,j}^d,...v_{i,m_i}^d\ ]$. In Figure \ref{p1}, $x_3 = ``\mbox{海(sea)''}$, for the first memory cell, we can map candidate word boundary information to the value list $\mathcal{V}_3^0\ = \ [V_S,V_B\ ]$. Four cells for $x_i$ has a values list $\mathcal{V}_i\ = \ [v_{i}^0,v_{i}^1,v_{i}^2,v_{i}^3\ ]$. Then the $d_{th}$ memory cell embedding for $x_i$ is computed from the weighted sum of all keys and values as follow.

\begin{equation}\label{eq8}
\textbf{o}_i^d\ =\ \sum_{j=1}^{m_i}{p_{i,j}^de_{i,j}^{v^d}}
\end{equation}
where $e_{i,j}^{v^d}$ is the embedding for $v_{i,j}^d$. Next, the final character embedding is the element-wise sum of $\textbf{o}_i$ and $\textbf{h}_i$, or their concatenation, passing through a fully connected layer as follow:
\begin{equation}\label{eq9}
\textbf{a}_i =\ \textbf{W}_o\cdot({\textbf{o}_i \odot \textbf{h}_i})
\end{equation}
where $\odot$ operation could be sum or concatenate, $\textbf{W}_o \in\ \mathbb{R}^T$ is a trainable parameter and the output $\textbf{a}_i\in\ \mathbb{R}^T$  is the final representation for the $i_{th}$ character. $\textbf{o}_i$ is the final memory embedding for the $i_{th}$ character, and can be calculated as follow.

\begin{equation}\label{eq10}
\textbf{o}_i= Switcher([\textbf{o}_i^0, \textbf{o}_i^1, \textbf{o}_i^2,\textbf {o}_i^3])
\end{equation}
where \emph{Switcher} is used to control how much information in each memory cell will be combined with the output of the encoder.

\subsubsection{Switcher} Inspired by the benefits of multi-task, a classifier has been added on top of the encoder to predict the era label of the input sentence. The discriminator predicts the probability of the correct era label, $z$, conditioned on the hidden states of the encoder, $\textbf{H}$, which is the output of ``[CLS]'' from BERT. The loss function of the discriminator is $\mathcal{J}_{disc}=\ -logP(z|\textbf{H})$, through minimizing the negative cross-entropy loss to maximizes $P(z|\textbf{H})$. 

In this study, $\textbf{H}$ is fed into a fully-connected layer and let it pass through a softmax layer to obtain probabilities for each era label.

\textbf{Switch mode.} For the switcher, we propose two switcher modes, \emph{ hard-switcher} and \emph{soft-switcher}.  \emph{Hard-switcher} switches memory cells according to the predicted final result from the discriminator. For the input sentence in Figure \ref{p1}, if the predicted result is the modern era, then the switcher will switch to the memory cell associated with modern Chinese, and ${\textbf{o}_{i\ }=\textbf{o}}_i^{d\ }$. \emph{Soft-switcher} switches memory cells according to the predicted probability, which is calculated by the weight of each memory cell. \emph{Soft-switcher} means that the information from all four dictionaries may be fused into the current character's representation. For example, the predicted probability list is $\ [0.1, 0.2, 0.1, 0.6\ ]$; therefore, the final memory representation for the $i_{th}$ character is ${\textbf{o}_{i\ }=\textbf{o}}_i^{0\ }\ \ast\ 0.1\ + \textbf{o}_i^{1\ }\ \ast\ 0.2\ + \textbf{o}_i^{2\ }\ \ast\ 0.1+ \textbf{o}_i^3\ \ast\ 0.6$.

\subsubsection{Objective} In this framework, the discriminator is optimized jointly with the CWS task, which both share the same encoding layer. Different weights are assigned to the loss of the two tasks, the final loss function is:

\begin{equation}\label{eq11}
\mathcal{J}={\ \alpha\mathcal{J}}_{cws}\ +(1\ -\ \alpha)\ \mathcal{J}_{disc}
\end{equation}
where $\alpha$ is the weight that controls the interaction of the two losses. $\mathcal{J}\ _{cws}$ is the negative log likelihood of true labels on the training set.

\begin{equation}\label{eq12}
\mathcal{J}\ _{cws}=\ -\sum_{n=1}^{N}{log(P\left(Y_n\middle| X_n\right))}
\end{equation}
where $N$ is the number of samples in the training set, and $Y_n$ is the ground truth tag sequence of the $n_{th}$ sample.

\begin{table*}[t]
\centering
\begin{tabular}{|c|c|c|c|c|c|c|c|c|} 
\hline
\multicolumn{3}{|c|}{Datasets}                             & Words & Chars & Word types & Char Types & Sents  & OOV Rate  \\ 
\hline
\multirow{6}{*}{ASACC}    & \multirow{2}{*}{AKIWI} & Train & 2.8M  & 3.2M  & 65.3K      & 7.5K       & 59.7K  & -         \\ 
\cline{3-9}
                          &                        & Test  & 0.2M  & 0.3M  & 15.7K      & 4.4K       & 5K     & 4.35\%    \\ 
\cline{2-9}
                          & \multirow{2}{*}{DKIWI} & Train & 2.2M  & 2.8M  & 44.3K      & 6.0K       & 50.1K  & -         \\ 
\cline{3-9}
                          &                        & Test  & 0.2M  & 0.3M  & 13.0K      & 3.8K       & 5K     & 4.91\%    \\ 
\cline{2-9}
                          & \multirow{2}{*}{PKIWI} & Train & 6.4M  & 7.8M  & 117.0K     & 7.2K       & 144.1K & -         \\ 
\cline{3-9}
                          &                        & Test  & 0.2M  & 0.3M  & 18.6K      & 4.4K       & 5K     & 1.71\%    \\ 
\hline
\multirow{2}{*}{SIGHAN05} & \multirow{2}{*}{MSR}  & Train & 2.4M  & 4.1M  & 88.1K      & 5.2K       & 86.9K  & -         \\ 
\cline{3-9}
                          &                        & Test  & 0.1M  & 0.2M  & 12.9K      & 2.8K       & 4.0K   & 2.60\%    \\
\hline
\end{tabular}
\caption{Detail of the four datasets.}
\label{t3}
\end{table*}

\section{Experiment}
\subsection{Datasets}
The model proposed in this study has been evaluated on four CWS datasets from Academia Sinica Ancient Chinese Corpus\footnote{\url{http://lingcorpus.iis.sinica.edu.tw/ancient}} (ASACC) and SIGHAN 2005 \citep{em:22}. The statistics of all the datasets are listed in Table~\ref{t3}. Among these datasets, PKIWI, DKIWI, AKIWI from ASACC, correspond to near ancient Chinese, middle ancient Chinese, ancient Chinese, respectively, and MSR from SIGHAN 2005 is a modern Chinese CWS dataset. It should be noted that PKIWI, DKIWI, and AKIWI are traditional Chinese and were translated into simplified Chinese prior to segmentation.

For PKIWI, DKIWI, and AKIWI, 5K examples were randomly picked as a test set; then, 10\% of examples were randomly selected from training set as the development set. Similar to previous work \citep{em:4}, all datasets are pre-processed by replacing Latin characters, digits, and punctuation with a unique token.

In the cross-era learning scenarios, all of the training data from four eras corpora were used as the training set. Then, all of the test data from four corpora were used as the cross-era test set to evaluate the model. Finally, F1 and OOV recall rates ($R_{oov}$) were computed according to the different eras.

\subsection{Experimental configurations}
In our experiments, for the encoder BERT, we follow the default setting of the BERT \cite{em:47}. The key embedding size and value embedding size are the same as the output of the encoder, and they have been randomly initialized. For the baseline model Bi-LSTM, the character embedding size is set to 300, and the hidden state is set to 100. For the transformer, the same settings as \citet{em:1} were followed. The loss weight coefficient $\alpha$ is a hyper-parameter that balances classification loss and segmentation loss; the model achieves the best performance when $\alpha$ is set to 0.7, which was identified by searching from 0 to 1 with the equal interval set to 0.1.

The words in the training set form the internal dictionary,
and high frequency bi-gram and tri-gram are extracted from each corpus as the external dictionary. Each dataset generates its own dictionaries. Finally, combine the external dictionary with the internal dictionary as the final dictionary.
\begin{table*}[t]
\centering
\setlength{\tabcolsep}{3mm}
\begin{tabular}{c|l|l|c|c|c|c|c} 
\Xhline{1.2pt}
\hline
\rowcolor{gray!25}
\multicolumn{1}{l}{\textbf{NO.}} & \textbf{En-De}                                        &     &\textbf{ AWIKI }                       & \textbf{PWIKI}                        & \textbf{DWIKI }                       & \textbf{MSR}                          & \textbf{Avg.   }                       \\ 
\hline
\multicolumn{8}{l}{\textbf{Single-era learning}}                                                                                                                                                                                              \\ 
\hline
\multirow{2}{*}{1}        & \multicolumn{1}{l|}{\multirow{2}{*}{BT-CRF}} & F   & \multicolumn{1}{r|}{97.62} & \multicolumn{1}{r|}{97.58} & \multicolumn{1}{r|}{97.19} & \multicolumn{1}{r|}{98.03} & \multicolumn{1}{r}{97.61}  \\ 
\cline{3-3}
                          & \multicolumn{1}{c|}{}                        &  $R_{oov}$ & \multicolumn{1}{r|}{68.85}  & \multicolumn{1}{r|}{76.58} & \multicolumn{1}{r|}{74.80} & \multicolumn{1}{r|}{\textbf{86.85} }& \multicolumn{1}{r}{76.77}  \\ 
\hline
\multicolumn{8}{l}{\textbf{Cross-era learning}}                                                                                                                                                                                                              \\ 
\hline
\multirow{2}{*}{2}        & \multirow{2}{*}{BL-CRF}                      & F   & 89.78                             &85.98                              &  87.04                            &93.81                              &   89.15                            \\ 
\cline{3-3}
                          &                                              &  $R_{oov}$ &45.55                              &  46.43                            &   37.51                           &  58.06                            &    46.89                           \\ 
\hline
\multirow{2}{*}{3}        & \multirow{2}{*}{BL-CRF+SM}                   & F   &90.66                              & 87.41                             &     89.18                         &  95.42                            & 90.66                              \\ 
\cline{3-3}
                          &                                              &  $R_{oov}$ & 43.48                             &  44.40                            &   32.78                           &    68.74                          & 47.35                              \\ 
\hline
\multirow{2}{*}{4}        & \multirow{2}{*}{TR-CRF}                      & F   &95.89                              &  95.43                            & 95.87                             &    92.68                          &       94.97                        \\ 
\cline{3-3}
                          &                                              &  $R_{oov}$ &57.87                              & 58.01                             & 47.07                             & 72.24                             &   58.80                            \\ 
\hline
\multirow{2}{*}{5}        & \multirow{2}{*}{TR-CRF+SM}                   & F   & 96.69                             &     97.04                         &        96.87                      &           96.71                   &       96.82                        \\ 
\cline{3-3}
                          &                                              &  $R_{oov}$ &     64.22                         &   57.23                           &        50.42                      &  71.34                            &  60.80                             \\ 
\hline
\multirow{2}{*}{6}        & \multirow{2}{*}{BT-CRF}                      & F    &   97.04     & 97.51    & 96.96       & 97.75    & 97.32 \\ 
\cline{3-3}
                          &                                              & $R_{oov}$ &68.78 &75.39 & 73.94 & 86.48  &76.15 \\ 
\hline
\multirow{2}{*}{7}        & \multirow{2}{*}{CROSSWISE}                   & F   & \textbf{98.46} & \textbf{98.04} &\textbf{98.42} & \textbf{98.04}& \textbf{98.24}  \\ 
\cline{3-3}
                          &                                              &  $R_{oov}$       &\textbf{83.88}       & \textbf{81.86}    &\textbf{77.25}    & 86.50  & \textbf{82.37}  \\
\hline
\Xhline{1.2pt}
\end{tabular}
\caption{Experimental results of the proposed model on the tests of four CWS datasets with different configurations. ``+SM'' indicates that the model uses the Switch-memory module. There are two blocks. The first block is results of the baseline model (BERT-CRF) on the single-era dataset. The second block consists of the results of cross-era learning model with different encoders (``BL'' for Bi-LSTM, ``TR'' for Transformer, ``BT'' for BERT, ``CROSSWISE'' for BERT-CRF+SM ). Here, F, $R_{oov}$ represent the F1 value and OOV recall rate respectively. The maximum F1 values are highlighted for each dataset.}
\label{t5}
\end{table*}

\begin{table*}[!ht]
\centering
\begin{tabular}{l|c|c|c|c|c|c|c|c} 
\Xhline{1.2pt}
\multicolumn{1}{c|}{\multirow{2}{*}{\textbf{Models}}} & \multicolumn{2}{c}{\textbf{AWIKI}}  & \multicolumn{2}{c}{\textbf{PWIKI}} & \multicolumn{2}{c}{\textbf{DWIKI}} & \multicolumn{2}{c}{\textbf{MSR}} \\ 
\cline{2-9}
\multicolumn{1}{c|}{}   & F     & {$R_{oov}$}    & F     & {$R_{oov}$}  & F     & {$R_{oov}$}  & F     & {$R_{oov}$}            \\
\hline

\citet{em:4}    &  -                &-      &   - & -    &   - & -     &   96.04 & 71.60                   \\ 
\hline
\citet{em:41}      &  - &    -    &   - & -     &   - & -     & 97.78                  & 64.20           \\

\hline
\citet{em:48}  &91.25  &   56.32   &   97.01   &48.09&  97.00 & 43.18     &   97.09 & 75.19               \\ 
\hline 
\citet{em:49}    &   -   &   -    &   - & -     &   - & -   & \textbf{98.40}                   & 84.87               \\ 

\hline
\citet{em:1}  &  96.44  &  65.06    &  95.83   & 63.75   &96.31     &  57.03    & 98.05                  & 78.92                \\ 
\hline  
\citet{em:2}    &  98.16&   78.97   & 97.70    &  75.69  &   98.12  & 74.28   & 98.29                  & 81.75                \\ 
\hline 
\citet{em:26} &   - &  -   &   -   &   -    &   - & -     &   98.28& \textbf{86.67}                \\ 
\hline 

CROSSWISE          &\textbf{98.46}&   \textbf{83.88}   &  \textbf{98.04}  & \textbf{81.86} & \textbf{98.42}     &   \textbf{77.25}    & 98.04                  &86.50           \\
\Xhline{1.2pt}
\end{tabular}
\caption{Performance (F1 value) comparison between CROSSWISE and  previous state-of-the-art models on the test sets of four datasets.}
\label{t6}
\end{table*}
\subsection{Overall results}
To begin, in this section, the experimental results of the proposed model on the test sets from the four cross-era CWS datasets are provided, which can be seen in Table \ref{t5}.
 
Several observations can be made from the data provided in Table \ref{t5}. First, BERT-CRF in single-era scenarios (ID:1 in Table \ref{t5}) and cross-era learning without the SM module (ID:6) are compared. As can be seen in the table, when mixing four datasets, the average F1 value of all datasets decreases slightly. Single-era dataset learning has an average F1 value of 97.61, while cross-era learning without the Switch-memory module has a 97.32 average F1 value. This indicates that performance cannot be improved by merely mixing several datasets. 
 
Second, the models with the SM mechanism (ID:3,5,7) outperformed the baseline models (ID:2,4,6) in terms of F1 value and $R_{oov}$ on all datasets. For example,  the average F1 score for BERT-CRF with SM module (ID:7) improved by 0.92\% when compared to BERT-CRF (ID:6), and the average $R_{oov}$ went from 76.15 to 82.37. This indicates that the Switch-memory can help improve segmentation and $R_{oov}$ performance by integrating era-specific knowledge.

Third, among different encoders, the improvement of the pre-trained encoder BERT on the F1 value is still significant. When using Bi-LSTM as the encoder (ID:2,3), the average F1 value and the $R_{oov}$ are 89.15 and 90.66, respectively. When using BERT as the encoder (ID:6,7), the F1 value improves by approximately 8\%. The reason for this may be that the pre-training processing supplements some effective external knowledge.


To further illustrate the validity and effectiveness of this model, the best results from this study are compared to works that have been previously identified as state-of-the-art. Various aspects of multi-domain and multi-criteria Chinese word segmentation are very similar to the tasks in this study; therefore, this study reproduced experiments on several previous word segmentation models using the four datasets identified in this research \citep{em:48,em:1,em:2}. For the multi-domain segmenter PKUSeg \citep{em:48}, four datasets were trained with the pre-trained mixed model. The comparison is shown in Table~\ref{t6}; the model from this study outperforms previous methods.

\begin{figure}[htbp]
\centering
\includegraphics*[clip=true,height=4.0cm,width=6cm]{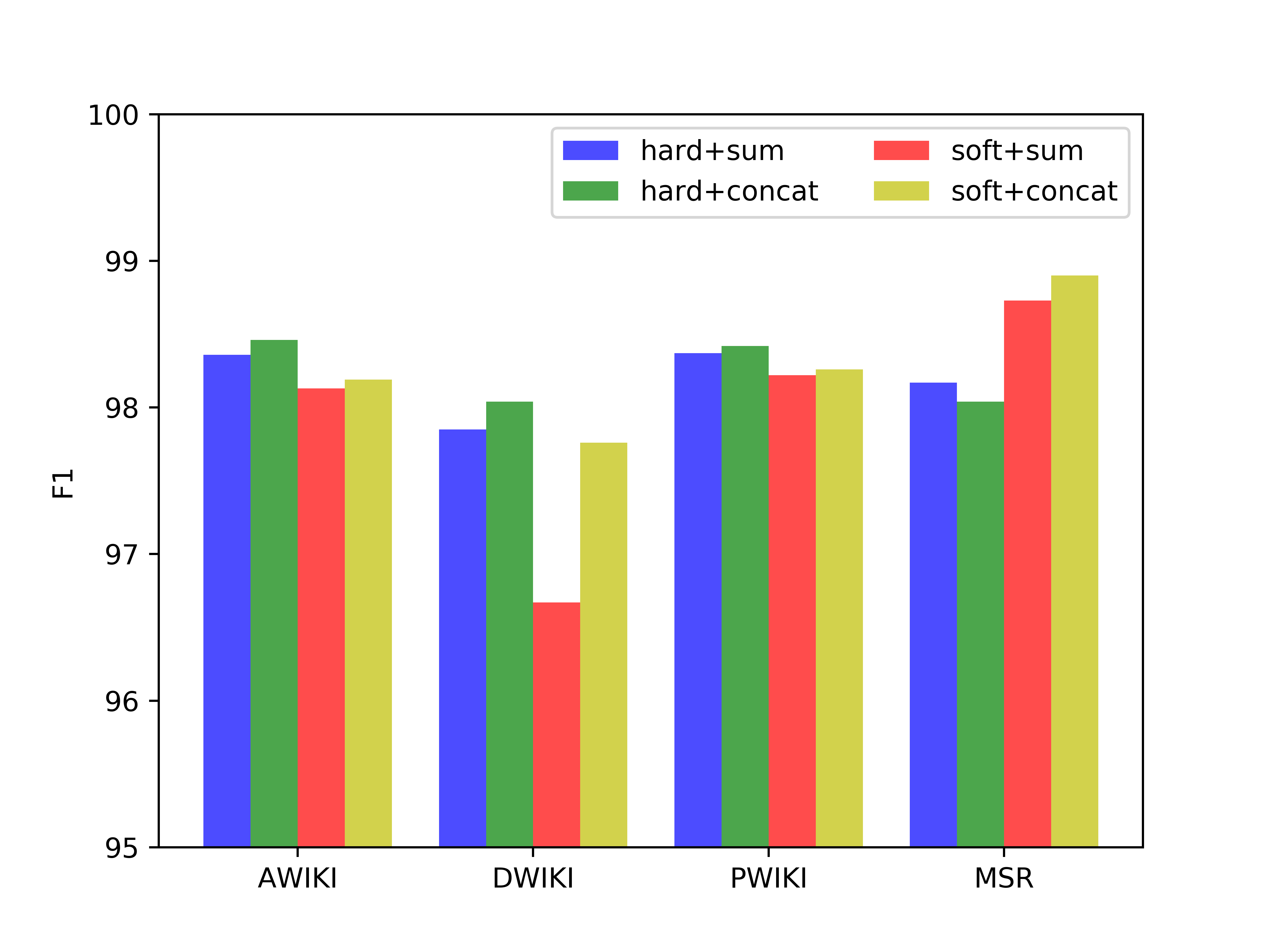}
\caption{ The F1 values of CROSSWISE using four pair settings; ``hard+sum'' means hard-switcher and the sum of the memory embedding and the character embedding from the encoder as the final character representation.}
\label{figure3}
\end{figure}

\begin{table*}[t]
\centering
\setlength{\tabcolsep}{1mm}
\begin{tabular}{|c|c|c|r|r|r|r|r|r|r|r|}
\hline
\multicolumn{1}{|c|}{\multirow{2}{*}{ID}} & \multicolumn{1}{c|}{\multirow{2}{*}{\textbf{Switcher}}} & \multicolumn{1}{c|}{\multirow{2}{*}{\textbf{Memory}}} & \multicolumn{2}{c|}{\textbf{AWIKI}}                         & \multicolumn{2}{c|}{\textbf{DWIKI}}                         & \multicolumn{2}{c|}{\textbf{PWIKI}}                         & \multicolumn{2}{c|}{\textbf{MSR}}                           \\ 
\cline{4-11}
\multicolumn{1}{|c|}{}                    & \multicolumn{1}{c|}{}                          & \multicolumn{1}{c|}{}                        & \multicolumn{1}{|c|}{F} & \multicolumn{1}{|c|}{$R_{oov}$} & \multicolumn{1}{|c|}{F} & \multicolumn{1}{|c|}{$R_{oov}$} & \multicolumn{1}{|c|}{F} & \multicolumn{1}{|c|}{$R_{oov}$} & \multicolumn{1}{|c|}{F} & \multicolumn{1}{|c|}{$R_{oov}$}  \\ 
\hline
1                                         &\checkmark                                        &  ×                                            & 98.00                     & 80.62                     & 97.87                  & 80.69                     & 97.52                  & 74.69                     & 98.01                  & 86.48                      \\ 
\hline
2                                         & ×                                               &\checkmark                                                & 98.28                  & 76.58                     & 97.85                  & 74.80                      & 98.32                  & 74.85                     & 98.63                  & \textbf{86.85}                      \\ 
\hline
3                                         &\checkmark                                                  &\checkmark                                                &  \textbf{98.46}                  & \textbf{83.88}                     & \textbf{98.04}                  & \textbf{81.86}                     & \textbf{98.42}                  & \textbf{77.25}                    & \textbf{98.04}                  & 86.50                      \\
\hline
\end{tabular}
\caption{Ablation experiments.}
\label{t7}
\end{table*}

\begin{table*}[!ht]
\centering
\begin{tabular}{|l|l|}
\hline
\rowcolor{gray!25}
\multicolumn{2}{|l|}{Sample from AWIKI (Ancient Chinese): 故上化下，犹风之靡草也。}\\
\rowcolor{gray!25}
\multicolumn{2}{|l|} {(Therefore, the superior civilizes and the subordinate, like the winds swept the grass) }        \\ 
\hline
\multirow{2}{*}{Golds} & 故/上/之/化/下/，/犹/风/之\textcolor[rgb]{1.00,0.00,0.00}{/靡/草/}也/。 
 \\ 
\cline{2-2}
 &So/superior/zhi/enlighten/subordinate/,/like/wind/zhi/swept/the/grass/. \\ 
\hline
w/o SM & 故/上/之/化/下/，/犹/风/之\textcolor[rgb]{1.00,0.00,0.00}{/靡草/}也/。 \\ 
\hline
Ours    & 故/上/之/化/下/，/犹/风/之\textcolor[rgb]{1.00,0.00,0.00}{/靡/草/}也/。\\ 
\hline
\rowcolor{gray!25}
\multicolumn{2}{|l|}{Sample from MSR (Modern Chinese): 天津市“鱼与熊掌兼得”的实践也就分外值得人们重视。}\\
\rowcolor{gray!25}
\multicolumn{2}{|l|} {(Tianjin's practice of ``getting both the fish and the paw'' deserves special attention.)}                        \\ 
\hline
\multirow{3}{*}{Golds} & 天津市/“\textcolor[rgb]{1.00,0.00,0.00}{/鱼/与/熊掌/}兼/得/”/的/实践/也/就/分外/值得/人们/重视/。 \\ 
\cline{2-2}
& Tianjin/``/fish/and/bear's paw/both/get/''/of/practice/also/then/extraordinary/\\&worth/people/important/.
\\ 
\hline
w/o SM & 天津市/“\textcolor[rgb]{1.00,0.00,0.00}{/鱼与熊掌/}兼/得/”/的/实践/也/就/分外/值得/人们/重视/。 \\ 
\hline
Ours    &天津市/“\textcolor[rgb]{1.00,0.00,0.00}{/鱼/与/熊掌/}兼/得/”/的/实践/也/就/分外/值得/人们/重视/。
  \\ 
\hline
\rowcolor{gray!25}
\multicolumn{2}{|l|}{Sample from MSR (Modern Chinese): 从大乱走向大治，中经雍正承前启后。} \\
\rowcolor{gray!25}
\multicolumn{2}{|l|}{(From chaos to prosperity, through Yongzheng connects the past and the future.)}       \\
\hline\multirow{2}{*}{Golds} &从/大/乱/走/向/大/治/，\textcolor[rgb]{1.00,0.00,0.00}{/中/经/}雍正/承前启后/。  \\ 
\cline{2-2}
 & From/big/chaos/go/to/big/prosperity/,/middle/through/Yongzheng/connect/.\\ 
\hline
w/o SM & 从/大/乱/走/向/大/治/，\textcolor[rgb]{1.00,0.00,0.00}{/中经/}雍正/承前启后/。  \\ 
\hline
Ours   & 从/大/乱/走/向/大/治/，\textcolor[rgb]{1.00,0.00,0.00}{/中/经/}雍正/承前启后/。\\
\hline
\end{tabular}
\caption{Segmentation cases from the test sets of MSR, AWKI and DWIKI datasets.}
\label{t9}
\end{table*}

\subsection{Ablation study}
Table~\ref{t7} shows the effectiveness of each component in the SM module.

The first ablation study is conducted to verify the effectiveness of memory cells. In this experiment, the sentence classification task is no longer a switcher but simply a joint training task with word segmentation. We can see that the ancient Chinese datasets (AWIKI, DWIKI, PWIKI) are more sensitive to memory cells than MSR. This may be explained by the fact that the encoder is pre-trained with a large quantity of modern Chinese data, and the memory cells in this study incorporate some ancient era knowledge into the model, which helps to boost the performance of the three ancient Chinese datasets.
  
The second ablation study is to evaluate the effect of the switcher. For this experiment, the average of four embedded memory cells is used as the final memory representation. The comparison between the second and the third line indicates that the switcher is an important component when integrating era-specific information. 

In summary, in terms of average performance, the switcher and the memory cells can both boost the performance of $R_{oov}$ considerably.

\subsection{Mode selection}

In this section, the effect of the switcher mode and the combination mode (concatenate or sum) of memory embedding and character embedding is investigated.

To better understand the effect of the different configurations, this study examines the four pair settings to train the model on the four datasets in this study; the results are shown in Figure \ref{figure3}, and different color bars represent different datasets. As can be seen, \emph{soft-switcher} significantly improves the F1 value on MSR compared to \emph{hard-switcher}, while the other three datasets prefer \emph{hard-switcher}, which suggests that the forward direction of knowledge dissemination from ancient Chinese to modern Chinese can help modern Chinese word segmentation, and that the reverse knowledge dissemination will have a negative impact on ancient Chinese word segmentation. Concatenating memory embedding and character embedding from the encoder outperforms the combination of the two; therefore, this study chose the pair of configurations, ``hard +concat'', to obtain the experimental results in the last row of Table \ref{t5} and Table \ref{t6}.


\subsection{Case study}
This study further explores the benefits of the SM mechanism by comparing some cases from BERT-CRF and CROSSWISE. Table~\ref{t9} lists three examples from the test sets of Ancient Chinese and modern Chinese datasets. According to the results, in the first sentence, ``{靡}(swept)'' and ``{草}(grass)'' are two words in ancient Chinese, BERT-CRF treats these two words as a single word; BERT-CRF gives the second sentence the wrong boundary prediction in ``{中}(middle)'' and ``{经}(through).'' However, this study’s CROSSWISE achieves all exact segmentation of these instances. The third sample is a sentence written in both ancient and modern Chinese,``鱼与熊掌兼得,'' which is a famous classical sentence in ancient Chinese. CROSSWISE also can split the sentence correctly. Therefore, it can be concluded that the model is flexible for Chinese word segmentation of era-hybrid texts and can produce different segmentation granularity of words according to the era of the sentence. Concurrently, it shows that the SM mechanism is effective in integrating era-specific linguistic knowledge according to different samples.

\section{Conclusion}
In this study, a flexible model, called CROSSWISE, for cross-era Chinese word segmentation is proposed. This model is capable of improving the performance of each dataset by fully integrating era-specific knowledge. Experiments on four corpora show the effectiveness of this model. In the future, the incorporation of other labeling tasks into CROSSWISE, such as POS tagging and named entity recognition, may prove to be insightful.

\section*{Acknowledgments} 
This research is supported by the NSFC project ``the Construction of the Knowledge Graph for the History of Chinese Confucianism'' (Grant No. 72010107003). We would like to thank Professor Jun Wang and Hao Yang for their insightful discussion.

\section*{Ethical Considerations}
The datasets used in this paper are open datasets and do not involve any ethical issues.
\bibliography{cite}
\bibliographystyle{acl_natbib}

\appendix

\end{CJK*}
\end{document}